# Domain Wall Magnetic Tunnel Junction Reliable Integrate and Fire Neuron

Can Cui[1,2], Sam Liu[1,2], Jaesuk Kwon[1,2], and Jean Anne C. Incorvia[1,2]

[1] *Chandra Family Department of Electrical and Computer Engineering, The University of Texas at Austin, Austin, TX, USA*

[2] *Microelectronics Research Center, The University of Texas at Austin, Austin, TX, USA*

**Abstract**

In spiking neural networks, neuron dynamics are described by the biologically realistic integrate-and-fire model that captures membrane potential accumulation and above-threshold firing behaviors. Among the hardware implementations of integrate-and-fire neuron devices, one important feature, reset, has been largely ignored. Here, we present the design and fabrication of a magnetic domain wall and magnetic tunnel junction based artificial integrate-and-fire neuron device that achieves reliable reset at the end of the integrate-fire cycle. We demonstrate the domain propagation in the domain wall racetrack (integration), reading using a magnetic tunnel junction (fire), and reset as the domain is ejected from the racetrack, showing the artificial neuron can be operated continuously over 100 integrate-fire-reset cycles. Both pulse amplitude and pulse number encoding is demonstrated. The device data is applied on an image classification task using a spiking neural network and shown to have comparable performance to an ideal leaky, integrate-and-fire neural network. These results achieve the first demonstration of reliable integrate-fire-reset in domain wall-magnetic tunnel junction-based neuron devices and shows the promise of spintronics for neuromorphic computing.

**Introduction**

Artificial neuron devices are the fundamental building blocks of neuromorphic hardware and are designed to mimic the structure and information processing of biological neurons. While past generations of neural networks have already incorporated the concept of neurons using the computational units such as McCollough-Pitts perceptron and activation function, the third-generation neural network[1], named spiking neural network (SNN), achieves a high degree of biological plausibility by capturing the asynchronous, massively parallel information processing of the neuron system where neurons communicate through current spikes as described by the integrate-fire (IF) model[2]. A neuron integrates current inputs from other neurons and builds up its membrane potential; once it exceeds the firing threshold, the neuron emits an output spike that is sent to downstream neural circuits. The IF neuron has been implemented in CMOS[3–5] as well as memristive[6–8] threshold switching devices[9,10].

On the other hand, spintronic devices are promising candidates for neuromorphic computing due to their high endurance[11], scalability[12], and CMOS compatibility[13]. The ability to electrically manipulate spin textures such as the domain wall (DW)[14–17] allows spintronic devices to intrinsically emulate both synapse[18–24] and neuron functions[25–30]. One such device is the three-terminal domain wall-magnetic tunnel junction (DW-MTJ) that leverages both the information carrier capability of the DW and the magnetization readout capability of the MTJ[18,31,32] to operate as either a synapse or neuron. As shown in Fig. 1a, the DW-MTJ device consists of an MTJ sitting on top of a DW racetrack. A DW initialized at the left end of the racetrack can be driven by electrical current through spin-transfer torque (STT) or spin-orbit torque (SOT) towards the MTJ to switch its free layer magnetization, which is detected by the magnetoresistance (MR) change of the MTJ. It can be seen that the DW motion and the MR switching of the MTJ mimic the membrane potential integration and firing event, respectively, of an IF spiking neuron. Such functionality has been experimentally shown for both in-plane and out-of-plane DW-MTJ devices[22,30,31].

However, one important aspect of the integrate-fire model – reset – has received little attention in the DW-MTJ IF neuron device. As in the biological neuron whose membrane potential recovers to the resting potential following the emission of current spike, the artificial IF neuron must also be reset to prepare for the next IF cycle. Volatile IF neurons[9] intrinsically achieve self-reset, while nonvolatile IF neurons often rely on peripheral CMOS circuits to perform reset. For the nonvolatile DW-MTJ neuron, the successful reset entails electrically driving the DW in the opposite direction back to the initial position, which in turns requires retaining the DW in the racetrack by engineering exchange-pinned "frozen spin" regions at both ends of the racetrack[33,34]. Such a reset scheme greatly increases the device design and manufacturing complexity and doubles the energy consumption per spike. It also creates a reliability challenge to contain the DW in the track back-and-forth over many cycles. Moreover, the DW-MTJ neuron and logic devices hitherto demonstrated either use an Oersted field line to initialize a DW[32] or use current-induced random nucleation of domains to leverage the stochastic switching of the device[30], therefore limiting the robust injection of a DW.

Here, we design and fabricate a DW-MTJ IF artificial neuron that achieves reliable reset and cycling performance using the motion of two, instead of one, DWs. We demonstrate the robust creation, propagation, readout, and reset of the domain over 100 cycles. The device data is shown to have comparable performance to an ideal leaky, integrate-and-fire spiking neural network for image classification. These results show that magnetic devices can capture the higher-order characteristics of biological neurons to create both reliable and efficient neural networks.

**Device Data Results**

The designed neuron device, as illustrated by the schematics in Fig. 1a, consists of two MTJ pillars with top-pinned reference layer and shared free layer in the shape of a nanotrack. Compared to the single MTJ version of the DW-MTJ device[18,30,32], the introduction of an additional MTJ has two purposes: a) to improve DW injection efficiency compared to using an Oersted field line or random defect DW nucleation, and b) to inject a reverse-magnetized domain bounded by two DWs. The device operation cycle consists of four steps as shown in Fig. 1b: a) write: a single voltage pulse is applied between $IN$ and $A$ terminals to nucleate a domain under $MTJ_A$; b) integrate: a train of voltage pulses is applied between the $IN$ and $GND$ terminals to depin the nucleated domain from $MTJ_A$ and drive it towards $MTJ_B$; c) fire: as the domain passes under $MTJ_B$, its MR is switched and detected through terminal $B$ and $GND$; and d) reset: as the domain is further driven forward, it is eventually depinned from $MTJ_B$ and driven out of the track.

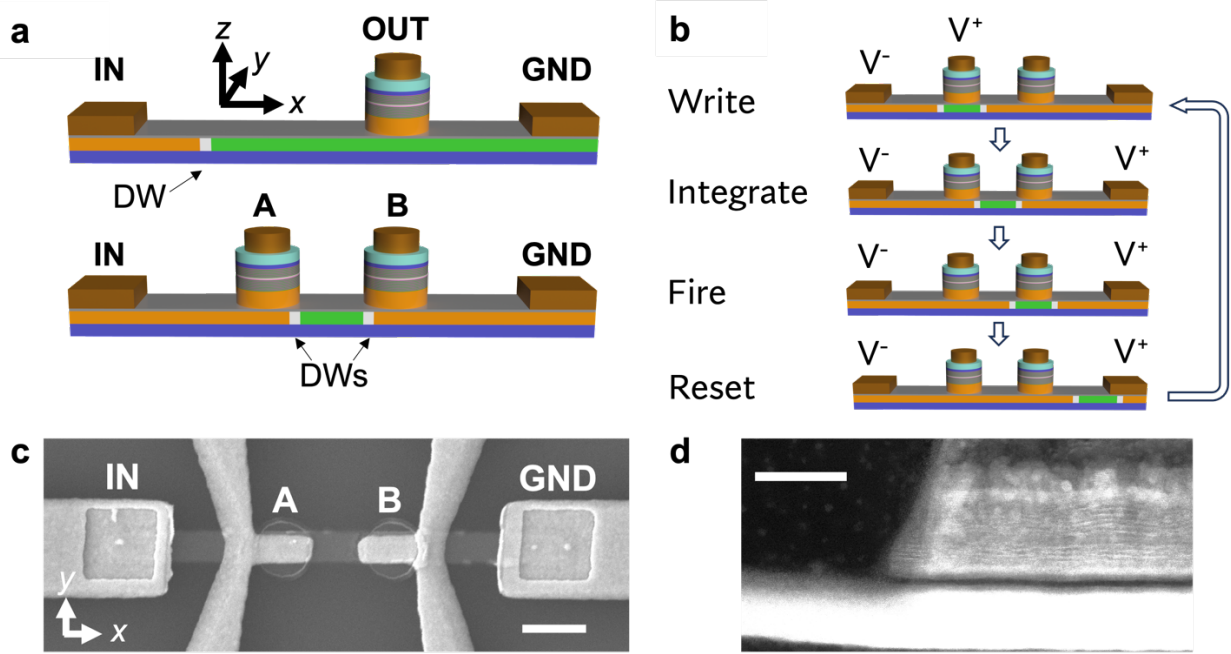

**Figure 1. Domain wall-magnetic tunnel junction (DW-MTJ) integrate-fire neuron with MTJs for write and read. a**, Three-terminal (3T, top) and four-terminal (4T, bottom) versions of DW-MTJ neurons, using one or two DWs as membrane potential integrator, respectively. Orange/green represent oppositely magnetized, perpendicular magnetic anisotropy domains, with white domain walls. Here, the 4T version is used with terminals IN, GND, A, and B. **b**, Illustration of the domain position during one integrate-fire-reset cycle of the 4T DW-MTJ neuron. **c**, SEM image of a fabricated device prototype. Scale: 500 nm. **d**, TEM image of the cross-section of the device after patterning, showing both the DW racetrack and MTJ$_B$. The MgO tunnel barrier is black with bright Ta/CoFeB DW racetrack at bottom. Scale: 20 nm.

The device prototype is fabricated from a spin transfer torque-magnetic random access memory (STT-MRAM) thin film multilayer stack with perpendicular magnetic anisotropy (PMA), grown by Applied Materials. The layers are Si/ SiO$_2$(1k)/ Ta(30)/ {CoFeB/Ta/CoFeB}(24)/ MgO(10)/ CoFeB(19)/ [Co/Pt](50)/ Ru(9)/ [Co/Pt](69)/ Ta(10)/ Ru(30) (thickness in Å); fabrication process is described in *Methods*. Figure 1c shows the scanning electron microscope (SEM) image of one fabricated device. The width of the DW track is 250 nm, and the dimensions of the MTJ pillars are 450 nm along the track length and 250 nm across the track. Figure 1d shows a transmission electron microscope (TEM) image of the Ta/CoFeB DW track (bottom, bright), MgO tunnel barrier (dark), and pinned layers, showing the careful etch stop at the MgO to preserve the free layer.

The measurement setup for the electrical characterization of the fabricated device is shown by Fig. 2a. The MTJs are first characterized by the magnetoresistance (MR) hysteresis loops with perpendicular magnetic field $H_z$ (Fig. 2b). It can be seen that both MTJs exhibit PMA. MTJ$_A$ has coercivity fields $H_{cA1} = -48\ Oe$, $H_{cA2} = 570\ Oe$; MTJ$_B$ has coercivities $H_{cB1} = -42\ Oe$, $H_{cB2} = 440\ Oe$. The asymmetric loops $|H_{c1}| \neq |H_{c2}|$ is due to the stray field along the $-z$ direction from the MTJ CoFeB pinned layers. Notably, the two MTJs have similar antiparallel to parallel (AP→P) switching field $H_{cA1} \approx H_{cB1}$; this is an important check to assure that the switching of the MTJs are dominated by domain nucleation and expansion in the DW track, confirming the free layer integrity. The P→AP switching field has a notable difference $|H_{cA2} - H_{cB2}| = 130\ Oe$. This is because of the difference in the pinned layers stray fields of the two MTJs that prevents the switching of the free layer magnetization from $-z$ to $+z$. The effect of such stray field difference is not evident in the AP→P switching, since in such case the field is assisting the switching of the free layer under the MTJs. The PMA of the DW track is further confirmed by measuring the anomalous Hall effect (AHE) resistance $R_{xy}$ of a Hall cross fabricated on the same chip as the neuron device. As shown in Fig. 2c, the $R_{xy} - H_z$ hysteresis loop of the Hall cross confirms the PMA of the CoFeB free layer. Coercivity difference between the AP→P switching and the P→AP is $|H_{c1} - H_{c2}| = 320\ Oe$, smaller than that of the MTJs, again suggesting the influence of MTJ CoFeB pinned layers stray field.

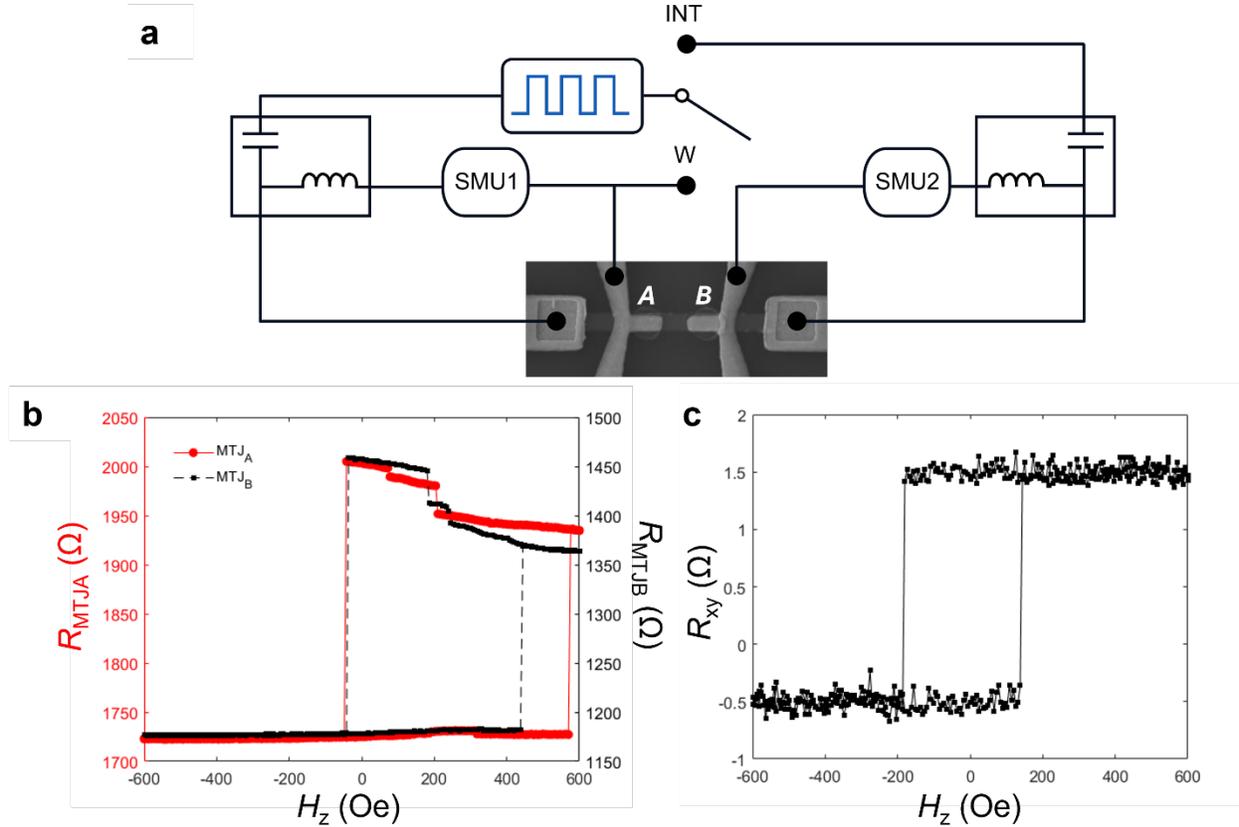

**Figure 2. Device electrical characterization setup and confirmation of free layer quality. a**, Device measurement setup with two source measurement units (SMUs) to simultaneously monitor magnetoresistance of the two MTJs, and a waveform generator (blue pulse train) to input voltage pulses between *IN* and *A* for DW write ("W") or integration ("INT"). **b**, Normalized out-of-plane field hysteresis loops of magnetoresistance of MTJ$_A$ (red solid) and MTJ$_B$ (black dashed). **c**, Out-of-plane field hysteresis loop of Hall resistivity $R_{xy}$ of a racetrack Hall cross fabricated on the same substrate as the DW-MTJ artificial neuron.

To utilize a magnetic domain in the racetrack as the membrane potential integrator, robust domain nucleation in the track is required. The measurement setup is shown in Fig. 2a. The switch is toggled to W ("Write") to apply voltage pulses through MTJ$_A$. Two identical source-measurement units (SMUs) are used to measure MR's of MTJ$_A$ and MTJ$_B$ simultaneously. First, an external field $H_z = -600\ Oe$ is applied to the device to saturate the free layer magnetization along $-z$, setting the MR of both MTJs to R$_P$ ($R_A = 1650\ \Omega$, $R_B = 1180\ \Omega$). Then, a voltage pulse with amplitude of 3.1 V is sent by the waveform generator through MTJ$_A$. All voltage pulse durations are set to $\tau = 50\ ns$ (including 10 $ns$ rise/fall time). A bias magnetic field $H_{bias} = +200\ Oe$ is applied along $+z$ to assist the P→AP switching of the MTJ. Note that the field is smaller than both the DW nucleation and propagation field, and therefore guarantees that no undesired random domain reversal occurs.

The MR versus pulse voltage ($R_{MTJ}$-V) for $MTJ_A$ and $MTJ_B$ are plotted in Fig. 3. Upon the application of the Write pulse, $R_A$ switches to 1930 Ω while $R_B$ remains unchanged. This indicates that a domain in $+z$ is locally nucleated in the DW track under $MTJ_A$. The measurement shown in Fig. 3a is performed after the domain is nucleated. Four states are marked out corresponding to four different combinations of $[R_P, R_{AP}]$ of both MTJs. State 1 ($V = 1.4\ V$) has $[R_A, R_B] = [R_{AP}, R_P]$, in which the domain is yet to depin from $MTJ_A$ after its nucleation. With more pulses with higher amplitudes, $[R_A, R_B] = [R_P, R_P]$ with $MTJ_A$ switched back to $R_P$ (state 2), indicating that the domain is now depinned from below $MTJ_A$ and driven into the track region between two MTJ pillars. Between $V = 2.0\ V$ and $V = 2.4\ V$, the domain "integrates" along the track in $+x$. At $V = 2.5\ V$, $MTJ_B$ switches to $R_{AP}$ and $[R_A, R_B] = [R_P, R_{AP}]$, meaning that the domain is now under $MTJ_B$, fully switching its MR (fire). Finally, with more pulses applied, at $V = 2.7\ V$, both MTJs switch back to $[R_A, R_B] = [R_P, R_P]$, fully resetting the device to the pre-write state. The device is now ready to input the next "Write" signal, without having to perform a hard reset step, such as driving a DW back-and-forth or re-saturation of the DW track with large external magnetic field.

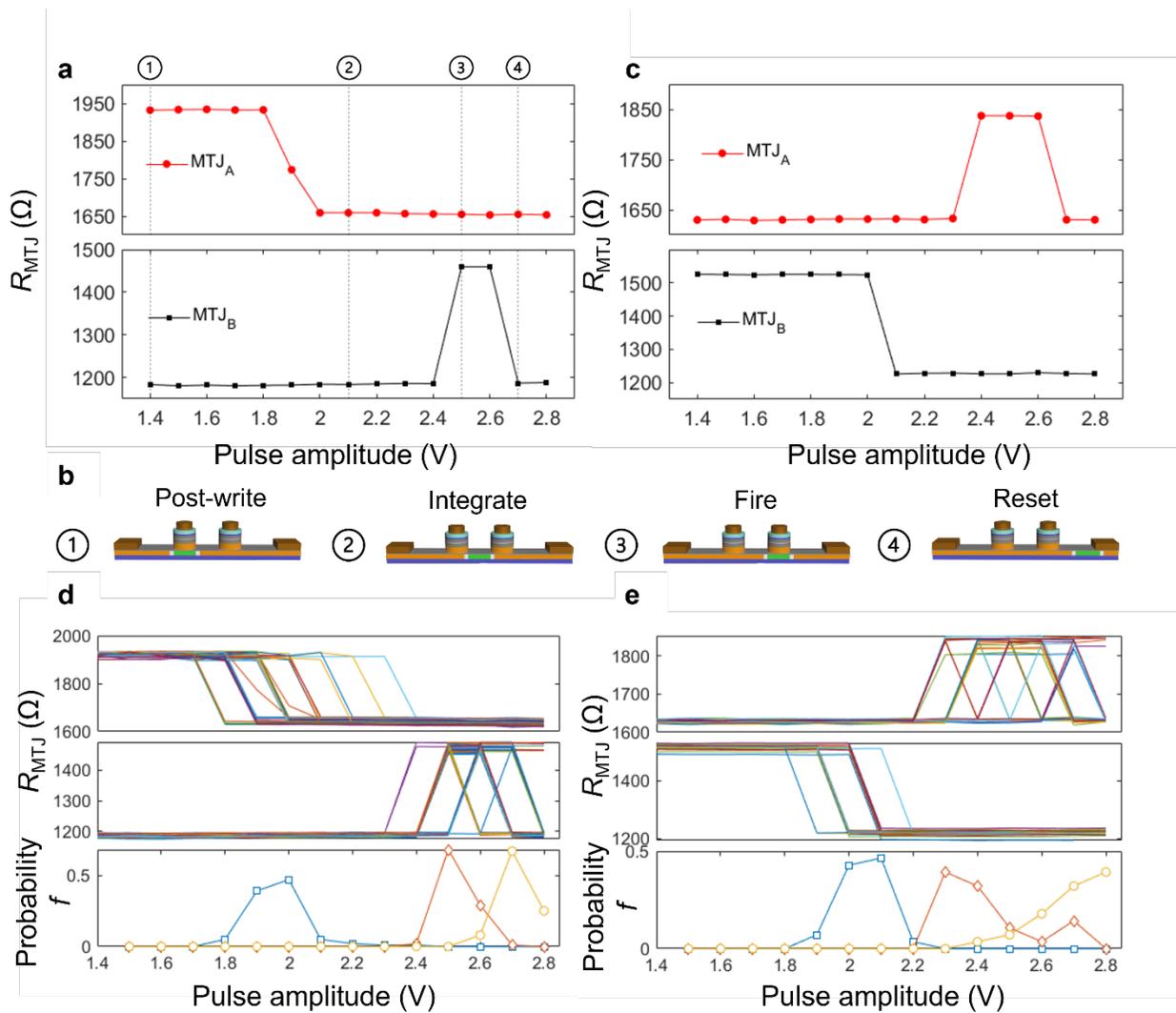

**Figure 3. Integrate-fire-reset cycle of the DW-MTJ neuron device. a**, Measured resistances of MTJ$_A$ (red circles) and MTJ$_B$ (black squares) versus pulse amplitude after the nucleation of the domain, with MTJ$_A$ as the Write MTJ and MTJ$_B$ as the Read MTJ. The resistance states are labeled and depicted in **b** showing (1) Write, (2) Integrate, (3) Fire, and (4) Reset. **c**, Same as a, but with MTJ$_B$ as the Write MTJ and MTJ$_A$ as the Read MTJ, showing the device can reliably operate in both directions. **d**, Data of 100 continuous write-integrate-fire-reset cycles from MTJ$_A$ to MTJ$_B$. **e**, 32 continuous write-integrate-fire-reset cycles from MTJ$_B$ to MTJ$_A$. The bottom panels show the probability of being in each state (blue squares = integrate, orange diamonds = fire, yellow circles = reset) over the cycles, showing separable regions in voltage amplitude for each state of the artificial neuron.

This data shows that the device demonstrates the integrate-fire-reset function as designed. To further confirm that the switching characteristics of the two MTJs are indeed due to the motion of the nucleated domain along the track, instead of random nucleation, the Write and Read MTJs are switched, i.e. the domain nucleation is performed through MTJ$_B$ and read out is performed through $MTJ_A$, and the above-described pulse experiment is repeated. As shown in Fig. 3c, similar switching characteristics are observed: the "fire-reset" event of the Read MTJ (in this case $MTJ_A$) follows the reset event of the Write MTJ ($MTJ_B$). The causality of the two events proves the directionality of the domain motion and therefore rules out random domain nucleation as MTJ switching mechanism. Moreover, the integrate-fire-reset behavior is robust with more than 100 cycles recorded, shown in Fig. 3d-e. It is worth noting that except for the 1$^{st}$ cycle, no field initialization is performed for subsequent integrate-fire-reset cycles, confirming the reliability of the device prototype. Probability $f$ shows a probability distribution function describing the switching probabilities of the data. The blue squares represent the probability that the domain will depin from $MTJ_A$ for a given voltage pulse amplitude, the orange diamonds represent the probability that the domain will move under $MTJ_B$, and the yellow circles represent the probability that the domain depins from under $MTJ_B$, after which the device is reset. The stochasticity of the DW behavior in the presented data is biomimetic and has been previously shown to aid in learning and inference tasks containing noisy data[30,35–37].

Data can be encoded for neural networks using numerous schemes, e.g. volage amplitude, pulse number, etc[38]. The above discussion uses voltage amplitude encoding: the pulses used to "integrate" the domain position are increased in voltage amplitude for each pulse. This scheme helps compensate for differences in depinning current along the track. In some devices the variation of DW depinning is minimized, allowing for the integrate-fire-reset to be observed with pulse number encoding using a constant-amplitude voltage pulse train, with an example shown in Fig. 4a. Here, the voltage pulse amplitude is fixed at 2.4 V and a pulse train of 20 pulses is applied to drive the motion. The first pulse pushes the domain out from under $MTJ_A$, starting integration. As the number of 2.4 V pulses increases, $MTJ_B$ fires and is then reset. Figure 4b shows 16 continuous cycles of the integrate-fire-reset behavior of one device to show reliable behavior.

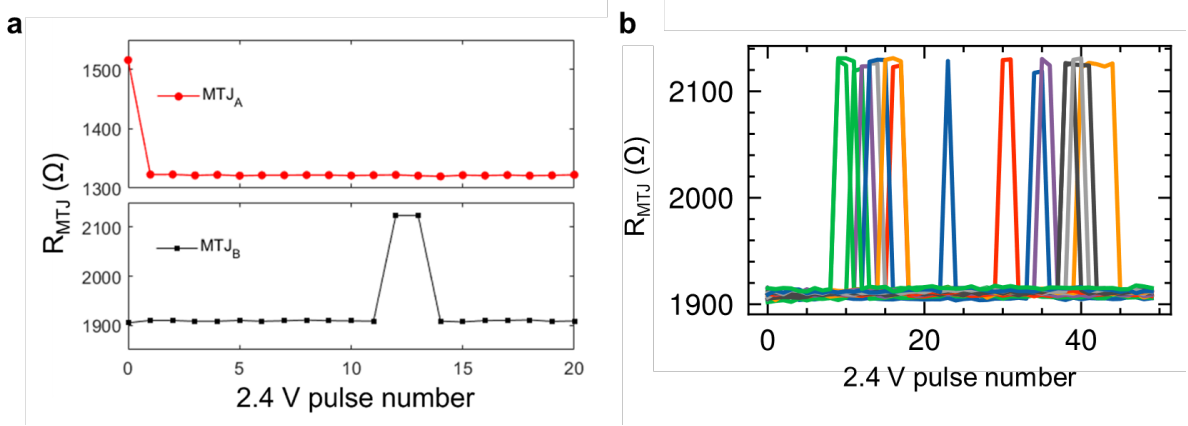

**Figure 4. Integrate-fire-reset cycle of the DW-MTJ neuron device with constant amplitude voltage pulses. a,** After the nucleation of the domain, measured resistances of MTJ$_A$ (red circles) and MTJ$_B$ (black squares) versus pulse number for 2.4 V, 50 ns pulses, with MTJ$_A$ as the Write MTJ and MTJ$_B$ as the Read MTJ. **b,** 16 continuous cycles of switching of MTJ$_B$.

**Neural Network Performance of Reliable IF DW-MTJs**

To directly connect the behavior of the device with analog implementations of neurons in a neural network, a PyTorch[39]-compatible model of the DW-MTJ neuron is constructed following the 1D solution of DW motion[40]. Since the data shows uniform motion of the two-DW domain across the device, it is valid to model as a 1D DW. The equation for the average DW velocity $\bar{v}$ is as follows:

$$\bar{v} = \frac{\gamma \Delta H_{eff}}{\alpha} + \frac{g\mu_B P}{2eM_{sat}}j$$

where $\gamma$ is the gyromagnetic ratio, $\Delta$ is the width of the DW, $H_{eff}$ is the effective external magnetic field, $g$ is the Lande factor, $\mu_B$ is the Bohr magneton, $e$ is electron charge, and $j$ is the applied current density applied in $-\hat{x}$. The cycle-to-cycle data in Figs. 3-4 shows there is stochasticity in the resistance state vs. pulse amplitude over cycles. To incorporate this observed variability within the model, variation is introduced to the change in DW position at each timestep $dt$ according to the following:

$$dx = \bar{v} * dt * N(1, \sigma)$$

Where $N(1, \sigma)$ is a random number drawn from a normal distribution with a mean of 1 and a set standard deviation $\sigma$. This model was fit to the experimental data in Fig. 4 with pulse number encoding. Figure 5a depicts the switching count of the experimental data shown in Fig. 4b, where switching count describes the number of runs where the DW takes the given number of pulses to move under $MTJ_B$, triggering a firing event. The aforementioned model was then fit to this data, where the value of $\sigma = 0.3$ can be understood to be approximately 30% variation in DW velocity where the variation is introduced for each timestep that $\bar{v}$ is calculated. The resulting distribution is shown in Fig. 5b, where the analytical model is applied with $\sigma = 0.3$. The blue set has an average switching threshold of 12 pulses and the green set has an average switching threshold of 35 pulses. Overall, the simulated data shows good agreement with the experimental data, validating its use in the subsequent neural network simulation.

Using this model based on the experimental data, a spiking multilayer perceptron shown in Fig. 5c was implemented in the Norse[41] SNN framework. The DW-MTJ neurons are implemented as the activations for the dense layers with ideal floating-point weights. The network is trained on the Fashion-MNIST[42] clothing article classification dataset with 60,000 images in the training set and 10,000 images in the test set. The test accuracy is shown for 10 epochs in Fig. 5d and compared with the same task when trained on a network with ideal leaky, integrate-and-fire (LIF) neurons shown in the dashed black line. The dataset shown in blue has deterministic DW dynamics of $\sigma = 0$, while the dataset shown in orange has the stochastic DW dynamics matching the experimental data with $\sigma = 0.3$. The shaded regions indicate the standard deviation across 6 runs for the ideal LIF (grey), deterministic DW-MTJ (blue), and stochastic DW-MTJ (orange). As shown here, when using DW-MTJ behavior, the average test accuracy is reduced by approximately 2%. Interestingly, the noisy DW-MTJ behavior results in slightly better accuracy than the deterministic DW-MTJ. Additionally, with the noisy DW-MTJ behavior, the standard deviation of the first three epochs is smaller than for the ideal LIF and deterministic DW-MTJ networks, at 1.1% for the noisy DW-MTJ, 2.2% for the deterministic DW-MTJ, and 2.4% for the LIF network. This implies that training with the DW-MTJ neurons results in more consistent initial training than for ideal LIF neurons. The advantage of having a noisy activation function allows the network to more fully explore the parameter space, which can improve the generalization of the SNN when the network is still actively learning[43,44], correlating with the improvement in consistency with the noisy DW-MTJ network. These results show that the DW-MTJ achieves comparable accuracy to the ideal LIF, while providing a reliable and compact formfactor for implementing the IF function. A comparison table in Supplementary Information benchmarks the results against other proposed IF artificial neurons, showing the DW-MTJ performs well under all three metrics of small area, high speed, and low energy efficiency.

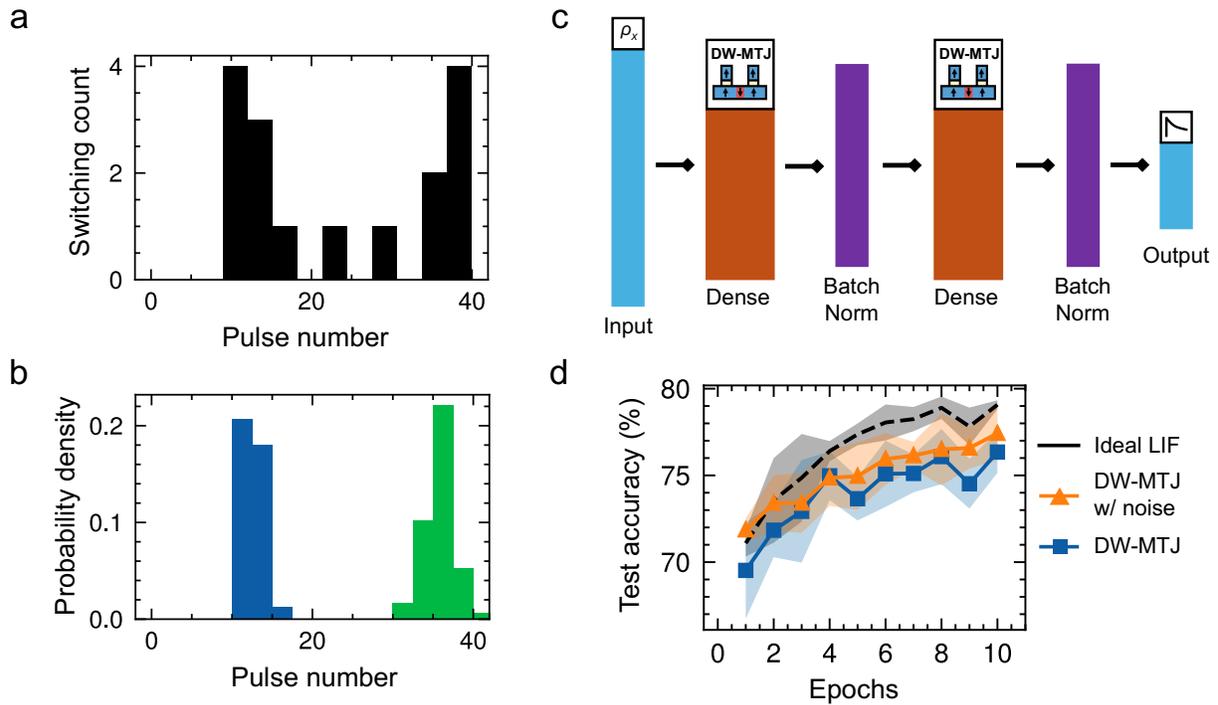

**Figure 5. Spiking neural network performance of the DW-MTJ IF neuron. a**, Histogram constructed from experimental data of constant voltage pulsing runs, where a switching count is recorded when the DW arrives under MTJ$_B$. **b**, Fitted switching probability density of 1D DW model with a pulsing threshold of 12 pulses (blue) and 35 pulses (green). **c**, Network architecture of multilayer perceptron constructed in Norse, where the DW-MTJ neurons are integrated at the end of the Dense layers. **d**, Test accuracy of Fashion-MNIST task comparing ideal LIF (black dashed), deterministic DW-MTJ (blue squares), and noisy DW-MTJ (orange triangles) networks.

**Conclusion**

We have designed and fabricated a novel integrate-and-fire artificial neuron device based on the domain wall-magnetic tunnel junction (DW-MTJ). The design presented in this work implements an MTJ for domain injection, allowing for reliable domain injection over cycles. The use of two, instead of one DW to emulate neuron membrane potential allows for the reset of the neuron after each operation cycle, simplifying the device design, reducing the energy consumption for device reset, and increasing cycling reliability. The experimental data fits well to a one-dimensional model of DW motion, which when implemented in a spiking multilayer perceptron shows classification accuracy approaching that of an ideal leaky-integrate-fire (LIF) neuron. Additionally, the stochastic DW-MTJ results indicate that the noisy spiking of magnetic systems helps improve the generalizability of the neuron during training. Combined with previous experimental demonstrations of DW-MTJ synapses, this work is an important step towards manufacturing reliable, large scale, all-spintronic neuromorphic systems.

**Acknowledgements**

This work was funded by the US National Science Foundation CAREER under award number 1940788, as well as funded, in part, from the US National Science Foundation Graduate Research Fellowship under award number 2021311125 (S.L.). J.K. acknowledges funding from Department of Energy Co-Design Project COINFLIPS. The fabrication work was done at the Texas Nanofabrication Facility supported by NSF Grant No. NNCI-1542159. The authors acknowledge the use of shared research facilities supported by the Texas Materials Institute.

**Data Availability**

All data needed to evaluate the conclusions in the paper are present in the paper. The code used to conduct NN simulations is available at https://github.com/liukts/autoreset-neuron.

**Methods**

The device prototype is fabricated from a spin-transfer torque-magnetoresistive random access memory (STT-MRAM) multilayer thin film stack grown by Applied Materials using the Endura Clover$^{TM}$ physical vapor deposition system. The domain wall racetrack and magnetic tunnel junctions are formed by electron beam lithography (Raith E-line) and ion milling (AJA). Silicon nitride dielectric is then deposited on top of the device by plasma-enhanced chemical vapor deposition (PECVD), and vias for making electrical contacts to the devices are formed by reactive ion etching (RIE). Finally, electrodes Cr(5)/Au(95) (in nm) are deposited using electron beam evaporation (Kurt J. Lesker PVD75).

DW-MTJ 1D Model Parameters:

| Parameter | Definition | Value |
|---|---|---|
| $M_{sat}$ | Saturation magnetization | 8E5 A/m |
| $P$ | Spin polarization | 0.7 |
| $\alpha$ | Damping constant | 0.05 |
| $\Delta$ | DW width | 9.7 nm |
| $w$ | DW track width | 25 nm |
| $d$ | DW track thickness | 1.5 nm |
| $dt$ | Timestep | 0.1 ns |

SNN Simulation Parameters:

| Parameter | Definition | Value |
|---|---|---|
| $\alpha_{LR}$ | Learning rate | 0.001 |
| $T$ | Sampling time | 40 ns |
| $f_{max}$ | Maximum Poisson frequency | $10^9$ |
| - | Batch size | 100 |